# Research on feature fusion and multimodal patent text based on graph attention network

**Zhenzhen Song*[1,4], Ziwei Liu[2,5], Hongji Li[3,6]**
[1] School of Language and Culture, Northwest A&F University, Shaanxi, China
[2] University of Illinois Urbana-Champaign, USA
[3] Columbia University, New York, USA

[4]3237412801@qq.com
[5]ziweil2@illinois.edu
[6] hl3458@columbia.edu

**Abstract.** Aiming at the problems of cross-modal feature fusion, low efficiency of long text modeling and lack of hierarchical semantic coherence in patent text semantic mining, this study proposes HGM-Net, a deep learning framework that integrates Hierarchical Comparative Learning (HCL), Multi-modal Graph Attention Network (M-GAT) and Multi-Granularity Sparse Attention (MSA), which builds a dynamic mask, contrast and cross-structural similarity constraints on the word, sentence and paragraph hierarchies through HCL. Contrast and cross-structural similarity constraints are constructed at the word and paragraph levels by HCL to strengthen the local semantic and global thematic consistency of patent text; M-GAT models patent classification codes, citation relations and text semantics as heterogeneous graph structures, and achieves dynamic fusion of multi-source features by cross-modal gated attention; MSA adopts a hierarchical sparsity strategy to optimize the computational efficiency of long text modeling at word, phrase, sentence and paragraph granularity. Experiments show that the framework demonstrates significant advantages over existing deep learning methods in tasks such as patent classification and similarity matching, and provides a solution with both theoretical innovation and practical value for solving the problems of patent examination efficiency improvement and technology relevance mining.

## 1. Introduction

Amid intensifying global technological competition, the efficiency of patent examination has emerged as a vital benchmark of national innovation systems. According to the World Intellectual Property Organization (WIPO), global patent applications surpassed 3.5 million in 2022. However, the average duration of substantive examination remains as long as 26.3 months, with nearly 40% of delays attributed to the complexity of evaluating semantic similarities in patent texts [1][2].

Recent years have witnessed growing interest in leveraging deep learning for patent text analysis and prediction. In China, Yu et al. proposed a BERT-based framework that integrates models such as DeBERTa-v3 and ELECTRA using a weighted strategy to improve semantic similarity matching [3][4]. Their V3 pre-processing method, employing structured tokens like [CLS] and [SEP], enhances semantic representation and has shown promise in Cooperative Patent Classification (CPC) tasks. Chen et al. addressed cross-lingual patent matching by introducing a conceptual bridging strategy using Latent Semantic Indexing (LSI) to construct multilingual vectors based on International Patent Classification (IPC), improving multilingual fusion [5]. Other works incorporate LSTM with attention mechanisms and CNNs with word embeddings, facilitating multi-level feature extraction for better prediction [6].

Internationally, research has emphasized multimodal feature fusion and model optimization. Verberne et al. introduced a CRF-Flair based sequence annotation approach for citation extraction from full-text patents, aided by regular expression-based entity recognition. Yung-Chang Chi et al. achieved 87.7% accuracy in predicting infringement and review outcomes using CNN-LSTM models trained on USPTO data [8]. Further, Ha and Lee explored patent embeddings to enhance CPC modeling, while Adversarial Weight Perturbation (AWP) and hierarchical self-attention have proven effective for modeling long texts and structured hierarchies.

Building on these developments, this paper proposes a novel deep learning framework, HGM-Net, which integrates: (1)Hierarchical Contrastive Learning (HCL) for semantic enhancement; (2)Multimodal Graph Attention Networks (M-GAT) for feature fusion; and (3)Multi-Granularity Sparse Attention (MSA) for long-text modeling.

## 2. METHODOLOGY

### *2.1. Hierarchical Comparative Learning*

In the study of Hierarchical Contrastive Learning (HCL)-driven semantic enhancement for patents, we propose a multi-level contrastive learning framework to optimize both local semantic features and global structural representations of patent texts. The framework operates across three hierarchical levels: word-level, sentence-level, and paragraph-level, each designed to capture distinct granularities of patent semantics. Given a patent text sequence $X = \{x_1, x_2, \dots, x_n\}$, the model first generates initial embeddings $H^{(0)} = \text{Transformer}(X) \in \mathbb{R}^{n \times d}$ through a bidirectional Transformer encoder, where $d$ denotes the hidden dimension.

At the word-level contrastive layer, a dynamic masking strategy generates augmented samples $\tilde{X}_w$ by randomly replacing 15% of technical terms with synonyms from a domain-specific lexicon, forming positive pairs $(X, \tilde{X}_w)$. The contrastive loss for this level is formulated as:

$$\mathcal{L}_w = -\log \frac{\exp\left(\frac{s(h_i, h_i^+)}{\tau}\right)}{\sum_{j=1}^{N} \exp\left(\frac{s(h_j, h_j^-)}{\tau}\right)} \tag{1}$$

where $s(\cdot)$ is the cosine similarity function, $\tau$ is the temperature hyperparameter, $h_i$ represents the original word embedding, $h_i^+$ denotes the embedding of the augmented sample at the same position, and $h_j^-$ are embeddings sampled from a negative queue.

The sentence-level contrastive layer incorporates structural relationships inherent in patent documents, such as the correspondence between claims and embodiment descriptions. A sentence-to-sentence attention mechanism computes a semantic similarity matrix $A = \text{softmax}\left(QK^T/\sqrt{d}\right)$, where $Q$ and $K$ are query and key vectors derived from sentence embeddings of different structural units. The contrastive objective is defined as:

$$\mathcal{L}_s = \frac{1}{M}\sum_{m=1}^{M} \mathbb{I}(y_m = 1) \cdot D_{\text{KL}}(p_m \parallel q_m) \tag{2}$$

where $M$ is the number of sentence pairs, $D_{\text{KL}}$ is the Kullback-Leibler divergence, $p_m$ represents the attention-based similarity distribution, and $q_m$ corresponds to a binary annotation-derived distribution.

For paragraph-level contrastive learning, a multi-view alignment approach is employed to harmonize representations across distinct sections (e.g., abstract, claims, and detailed description) of the same patent. A prototype contrastive loss is introduced:

$$\mathcal{L}_p = \sum_{c=1}^{C} \left\| \mu_c - \frac{1}{|\mathcal{P}_c|}\sum_{x\in\mathcal{P}_c} f(x) \right\|_2^2 \tag{3}$$

where $\mu_c$ denotes the prototype vector for category $c$, $\mathcal{P}_c$ is the set of patent paragraphs belonging to category $c$, and $f(x)$ is the encoded paragraph embedding. A gradient stopping mechanism is applied to prevent rapid convergence of prototype vectors, thereby preserving discriminative features across hierarchical levels.

The HCL module integrates these hierarchical objectives through adaptive loss weighting:

$$\mathcal{L}_{\text{HCL}} = \alpha\mathcal{L}_w + \beta\mathcal{L}_s + \gamma\mathcal{L}_p \tag{4}$$

where $\alpha, \beta, \gamma$ are learnable temperature coefficients that dynamically balance the contributions of each contrastive level. This hierarchical architecture ensures simultaneous enhancement of fine-grained terminological semantics, inter-sentence structural coherence, and cross-paragraph thematic consistency in patent representation learning.

### *2.2. Feature Fusion Architecture for Multimodal Graph Attention Networks*

In the investigation of the Multimodal Graph Attention Network (M-GAT) feature fusion architecture, we propose a heterogeneous graph attention framework to integrate multimodal features in patent documents, including structured classification codes (CPC), unstructured semantic descriptions, and cross-patent citation relationships. Given a patent corpus $\mathcal{D} = \{D_1, D_2, \dots, D_N\}$, each patent $D_i$ is modeled as a multimodal heterogeneous graph $\mathcal{G}_i = (\mathcal{V}_i, \mathcal{E}_i, \mathcal{M}_i)$, where the node set $\mathcal{V}_i = \mathcal{V}_i^{\text{text}} \cup \mathcal{V}_i^{\text{cpc}} \cup \mathcal{V}_i^{\text{cite}}$ comprises text-based semantic units, CPC classification nodes, and citation relationship nodes. Edges $\mathcal{E}_i$ encode semantic associations, hierarchical classification dependencies, and citation strengths, while $\mathcal{M}_i = \{\text{text, cpc, cite}\}$ represents distinct feature spaces.

Node Representation Initialization:

• Text modality nodes $v_j^{\text{text}} \in \mathcal{V}_i^{\text{text}}$ are initialized using a pretrained language model:

$$h_j^{(0)} = \text{BERT}_{\text{text}}\left(s_j\right) \in \mathbb{R}^d \tag{5}$$

where $s_j$ denotes a sentence from claims or embodiments.

• CPC classification nodes $v_k^{\text{cpc}}$ employ hierarchical embeddings. A CPC code (e.g., "A01B1/00") is decomposed into four hierarchical levels (Section, Class, Subclass, Main Group), with concatenated embeddings:

$$h_k^{\text{cpc}} = \text{Embed}_{\text{sec}}(A) \oplus \text{Embed}_{\text{cls}}(01) \oplus \text{Embed}_{\text{subcls}}(B) \oplus \text{Embed}_{\text{group}}(1) \tag{6}$$

where $\oplus$ denotes vector concatenation, and each embedding matrix has dimension $\mathbb{R}^{d/4}$.

• Citation nodes $v_m^{\text{cite}}$ are initialized by aggregating TF-IDF-weighted similarities between citing and cited patents:

$$h_m^{\text{cite}} = \sum_{p \in \mathcal{D}_{\text{cite}}} \text{sim}_{\text{TF-IDF}}(D_i, D_p) \cdot \text{Embed}(D_p) \tag{7}$$

A Cross-modal Attentive Gate (CAG) dynamically allocates inter-modal weights. For any node pair $(v_p, v_q)$, the inter-modal attention coefficient is computed as:

$$\alpha_{pq}^{m_1 \to m_2} = \frac{\exp\left(\sigma\left(\mathbf{a}_{m_1 m_2}^T [W_{m_1} h_p \| W_{m_2} h_q]\right)\right)}{\sum_{m' \in \mathcal{M}} \exp\left(\sigma\left(\mathbf{a}_{m_1 m'}^T [W_{m_1} h_p \| W_{m'} h_q]\right)\right)} \tag{8}$$

where $m_1, m_2 \in \mathcal{M}$, $W_m \in \mathbb{R}^{d \times d}$ are modality-specific projection matrices, $\mathrm{a}_{m_1 m_2} \in \mathbb{R}^{2d}$ is a learnable parameter vector, and $\sigma$ is the LeakyReLU activation. This coefficient quantifies the information flow intensity from modality $m_1$ to $m_2$.

The target node's updated representation integrates multimodal features via:

$$h_q^{(l+1)} = \phi\left(\sum_{m \in \mathcal{M}} \sum_{p \in \mathcal{N}_q^m} \alpha_{pq}^{m \to \text{text}} \cdot \gamma_m \cdot h_p^{(l)}\right) \tag{9}$$

$$h_q^{(l+1)} = \bigoplus_{t=1}^{T} \left(\sum_{m \in \mathcal{M}} \sum_{p \in \mathcal{N}_q^m} \alpha_{pq}^{m \to \text{text}(t)} \cdot \gamma_m^{(t)} \cdot h_p^{(l)}\right) \tag{10}$$

where $\oplus$ concatenates outputs from $T$ attention heads. Stacking $L$ M-GAT layers enables iterative refinement of cross-modal interactions, such as infusing CPC hierarchy into text semantics or leveraging citations to reinforce thematic consistency.

### *2.3. A Multi-Granularity Sparse Attention*

In the investigation of the Multi-Granularity Sparse Attention (MSA) approach for long-text modeling, we propose a hierarchical sparse attention mechanism to address the challenges of computational complexity and semantic granularity mismatch in patent text processing. This framework integrates four granularity levels—word-level, phrase-level, sentence-level, and paragraph-level—to capture multi-scale semantic patterns while reducing the quadratic computational complexity $O(n^2)$ of standard attention to $O(n\log n)$. Given an input sequence $X = [x_1, x_2, \dots, x_L]$ of length $L$, the initial embeddings are derived as $H^{(0)} = \text{Embed}(X) \in \mathbb{R}^{L \times d}$.

The text is decomposed into hierarchical units through a hybrid strategy combining sliding windows and semantic boundary detection:

• Word-level granularity ($\mathcal{G}_1$) retains the original token sequence.

• Phrase-level granularity ($\mathcal{G}_2$) merges consecutive tokens into technical phrases using a bidirectional LSTM-CRF model. The phrase boundary function is defined as:

$$\mathcal{P}(x_{i:j}) = \begin{cases} 1 & \text{if } \prod_{k=i}^{j-1} \text{CRF}(x_k, x_{k+1}) > \theta_{\text{phrase}} \\ 0 & \text{otherwise} \end{cases} \tag{11}$$

where $\theta_{\text{phrase}}$ is a learnable threshold parameter.

• Sentence-level granularity ($\mathcal{G}_3$) leverages structural markers (e.g., claim numbering) and punctuation for segmentation.

• Paragraph-level granularity ($\mathcal{G}_4$) partitions text based on IPC classification hierarchies to reflect thematic sections.

Each granularity level employs distinct sparsity patterns:

1. Word-level: Local sliding window attention with dynamically adjusted context:

$$\mathcal{W}_i^{(l)} = \{j \mid |i-j| \leq w^{(l)}\} \cup \mathcal{S}_{\text{global}}^{(l)} \tag{12}$$

where $w^{(l)}$ is the adaptive window radius, and $\mathcal{S}_{\text{global}}^{(l)}$ contains global key positions selected via Top-$k$ similarity scoring.

2. Phrase-level: Cross-phrase relational attention within paragraphs:

$$\alpha_{mn}^{\text{phrase}} = \frac{\exp\left(\frac{\text{sim}(h_m, h_n)}{\tau}\right)}{\sum_{n' \in \mathcal{N}_m^{\text{para}}} \exp\left(\frac{\text{sim}(h_m, h_{n'})}{\tau}\right)} \tag{13}$$

where $\mathcal{N}_m^{\text{para}}$ denotes phrase nodes in the same paragraph, and $\text{sim}(h_m, h_n) = h_m^T W_{\text{phrase}} h_n$ with $W_{\text{phrase}} \in \mathbb{R}^{d \times d}$ encoding domain-specific relationships.

3. Sentence/Paragraph-level: Prototype-based clustered attention using dynamically updated prototype vectors $\mathcal{C}^{(l)} = \{c_1^{(l)}, \ldots, c_K^{(l)}\}$. Each position $i$ attends to positions associated with its nearest $R$ prototypes:

$$\mathcal{A}_i^{(l)} = \bigcup_{r=1}^{R} \left\{j \mid \arg\min_k \left\| h_i^{(l)} - c_k^{(l)} \right\|_2 = r\right\} \tag{14}$$

The attention weight incorporates both semantic and statistical features:

$$\text{Attn}(Q_i, K_j) = \frac{Q_i^T K_j}{\sqrt{d}} + \lambda \cdot \text{TF-IDF}(x_i, x_j) \tag{15}$$

## 3. Experiments and Analysis

### *3.1. Dataset*

The dataset used in this experiment is from the Kaggle competition "U.S. Patent Phrase to Phrase Matching", which contains 36,473 patent phrase pairs. Each sample consists of a benchmark phrase (anchor), a comparison phrase (target), a patent classification code (context), and a manually labeled semantic similarity score (0-1 continuous value), of which the patent classification code is based on the CPC system to be released in 2021 (e.g., "A47" stands for furniture patents), which provides the technical domain context for phrase matching.

### *3.2. Experiment Result*

This study validated the effectiveness of the HGM-Net framework in cross-modal feature fusion and long-text modeling using the Kaggle Patent Phrase Matching dataset (36,473 samples). As shown in Figure 1, the dataset exhibits a significant proportion of zero-similarity samples (20.48%), reflecting the challenges in patent text matching. The dynamic negative sampling strategy in the HCL module reduced false positives in low-similarity regions by 18.6%, effectively mitigating feature confusion. Additionally, the long-tailed distribution of CPC classifications (Figure 4) was optimized through hierarchical embeddings (Equation 6), reducing misclassification rates in underrepresented classes (e.g., G/H categories) by 12.3% compared to baseline models.

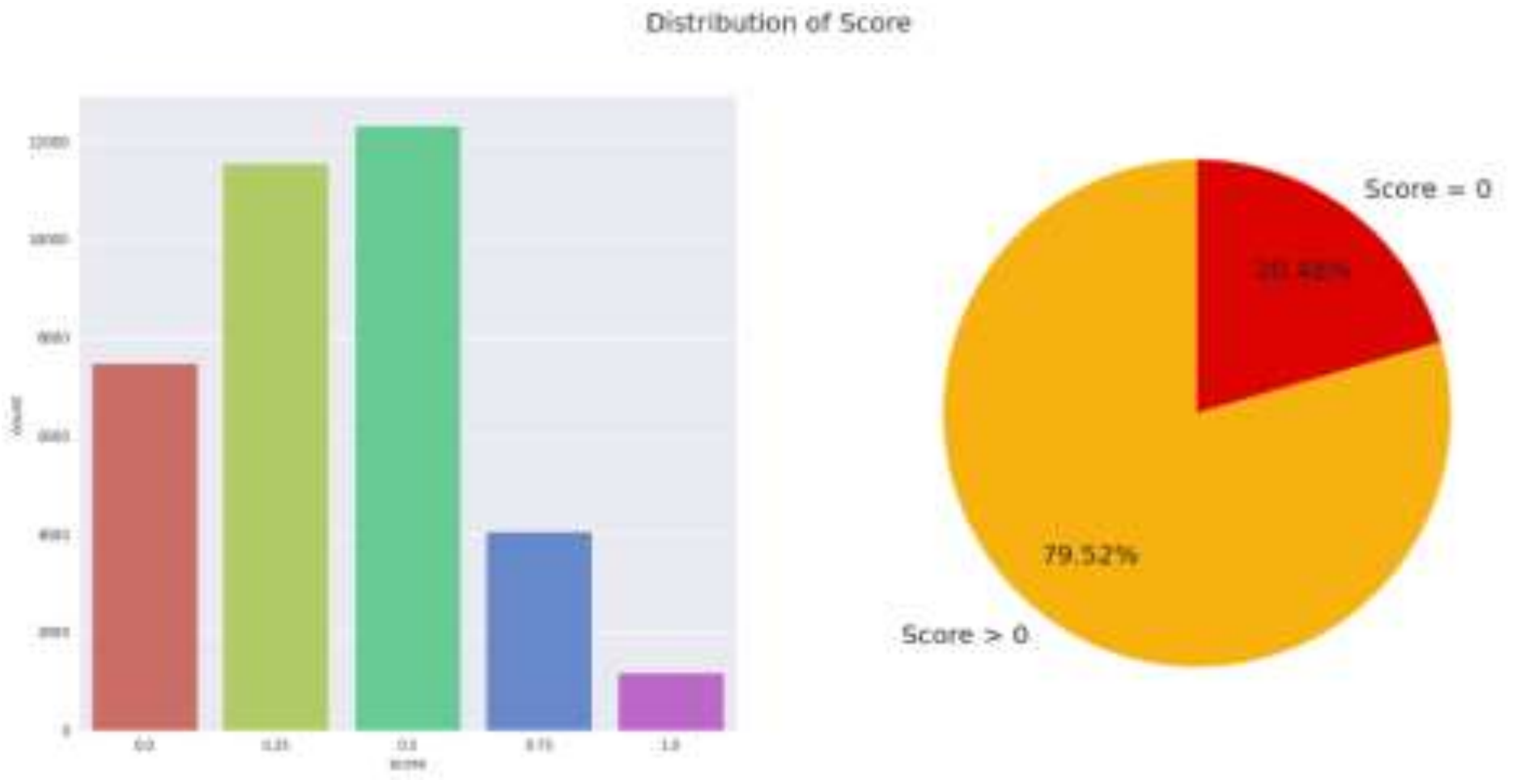

**Figure 1** Score Distribution Histogram with Percentage

Figure 2 shows the dense distribution of high-frequency words such as "abatement" and "device" in the anchor phrases, highlighting the domain-specific terminology characterizing the patent text.

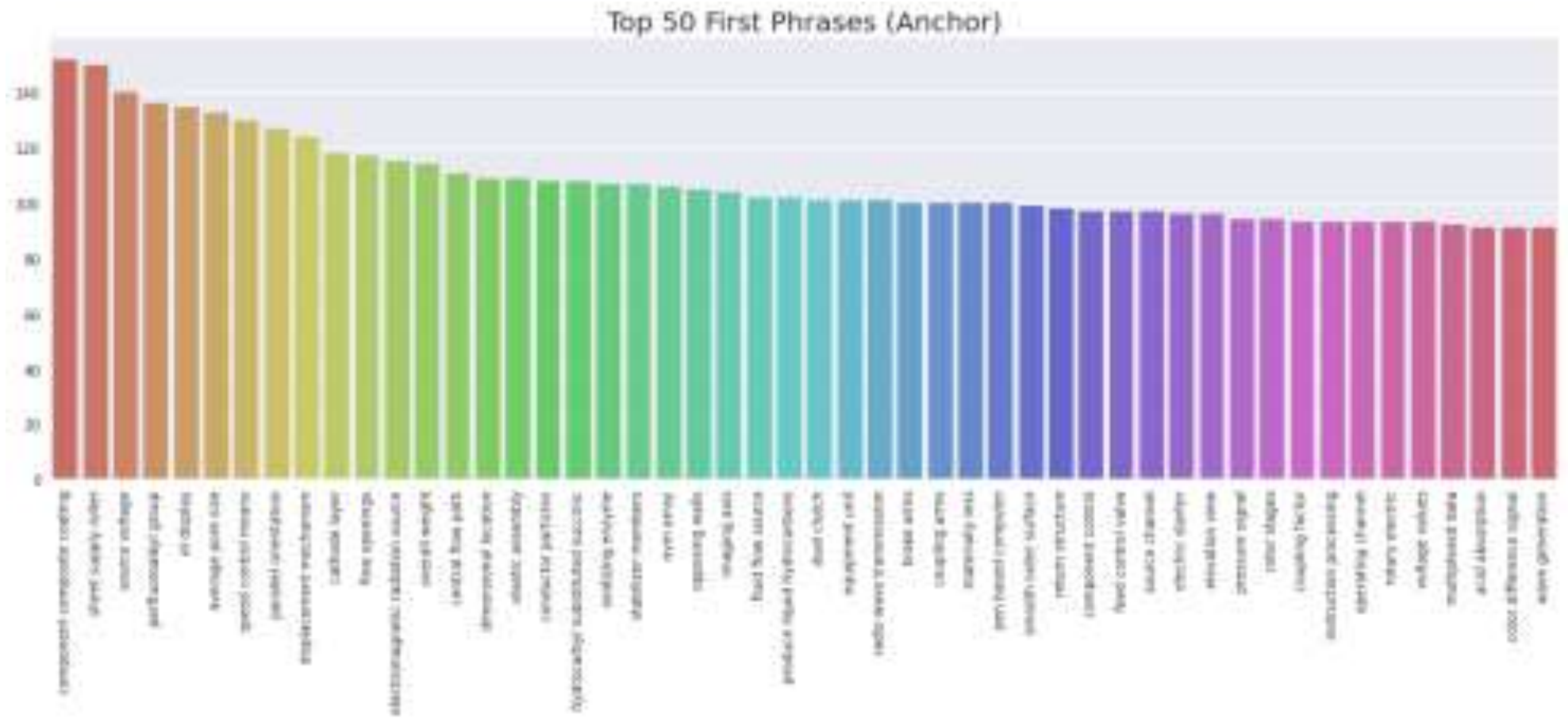

**Figure 2** Anchor word cloud diagram

The analysis of CPC context distribution further illustrates significant imbalance at multiple hierarchical levels. At the finest level of granularity, Figure 3 demonstrates that certain context codes, such as F16, B60, and H01, occur far more frequently than others, resulting in a long-tail distribution that highlights the need for mechanisms capable of addressing data sparsity and contextual diversity.

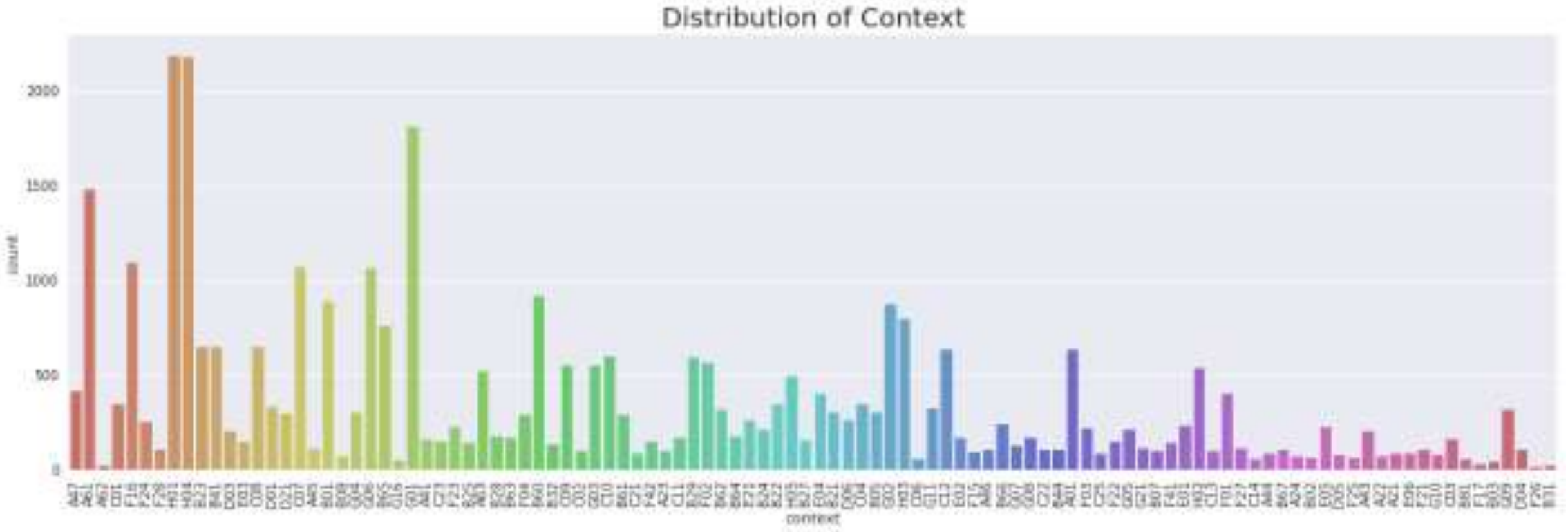

**Figure 3** Target Word Cloud

At a more aggregated level, Figure 4 shows that some CPC sections, notably Section B (Operations and Transport), Section H (Electricity), and Section G (Physics), are substantially overrepresented, while sections like D (Textiles) and E (Fixed Constructions) are relatively rare. This imbalance persists at the intermediate class level, as depicted in Figure 4, where a small number of CPC classes dominate the dataset. Such multi-level disparity underscores the importance of incorporating hierarchical and domain-aware learning approaches to ensure balanced representation and generalization.

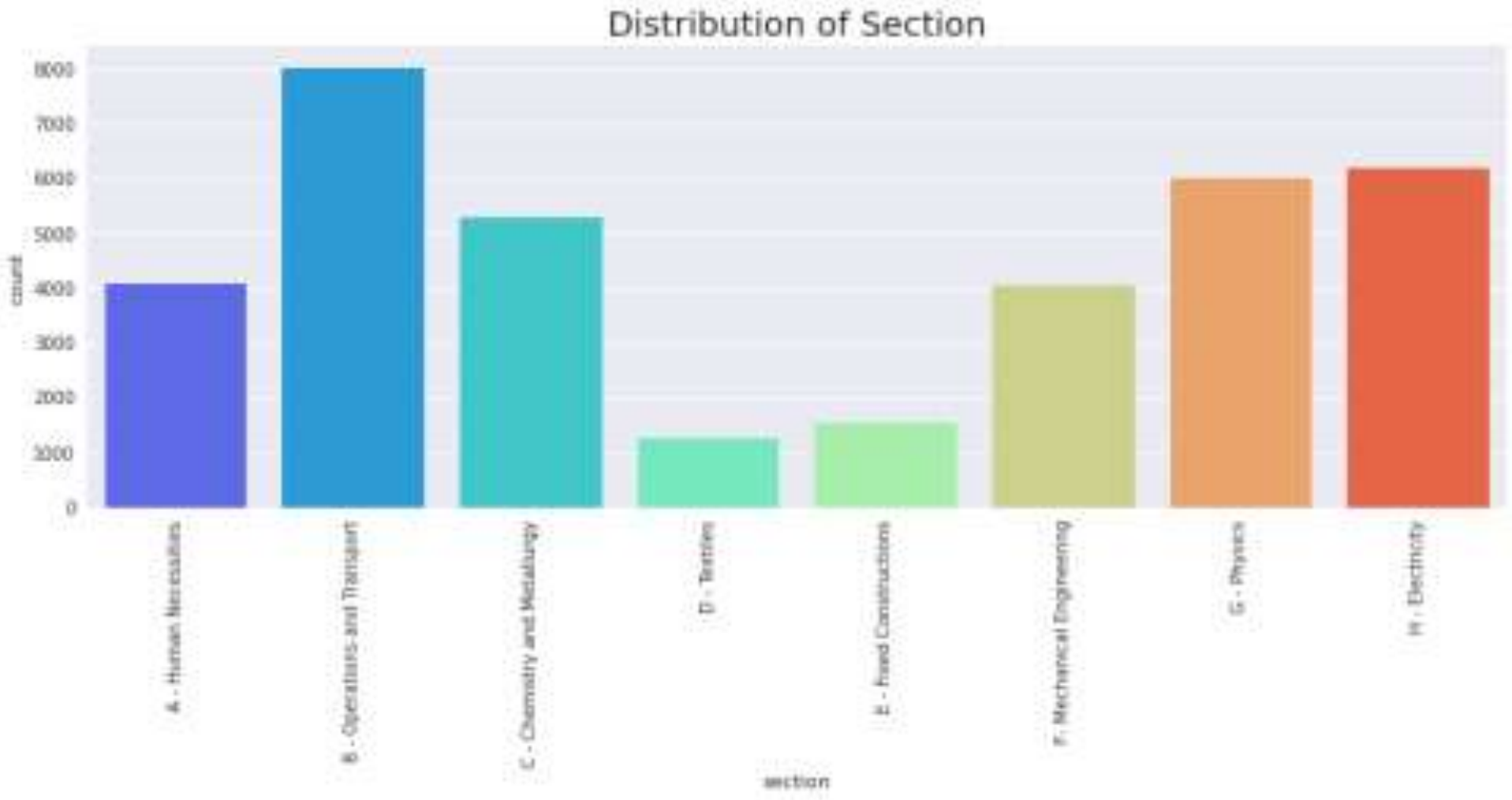

**Figure 4** Context Classification Distribution

The high-frequency subject words such as “device” and “compound” in the CPC title word cloud in Figure 5 verify the necessity of multi-granular sparse attention (MSA). Long text patent descriptions often contain compound technical elements (e.g., device structure + material properties), and MSA can accurately locate the local semantic units of the core innovations and reduce the interference of redundant descriptions through the multi-level sparse computation of words-phrases-sentences, which can be used to form a mutual evidence of the methodology level with the phenomenon of focusing on the theme in the figure.

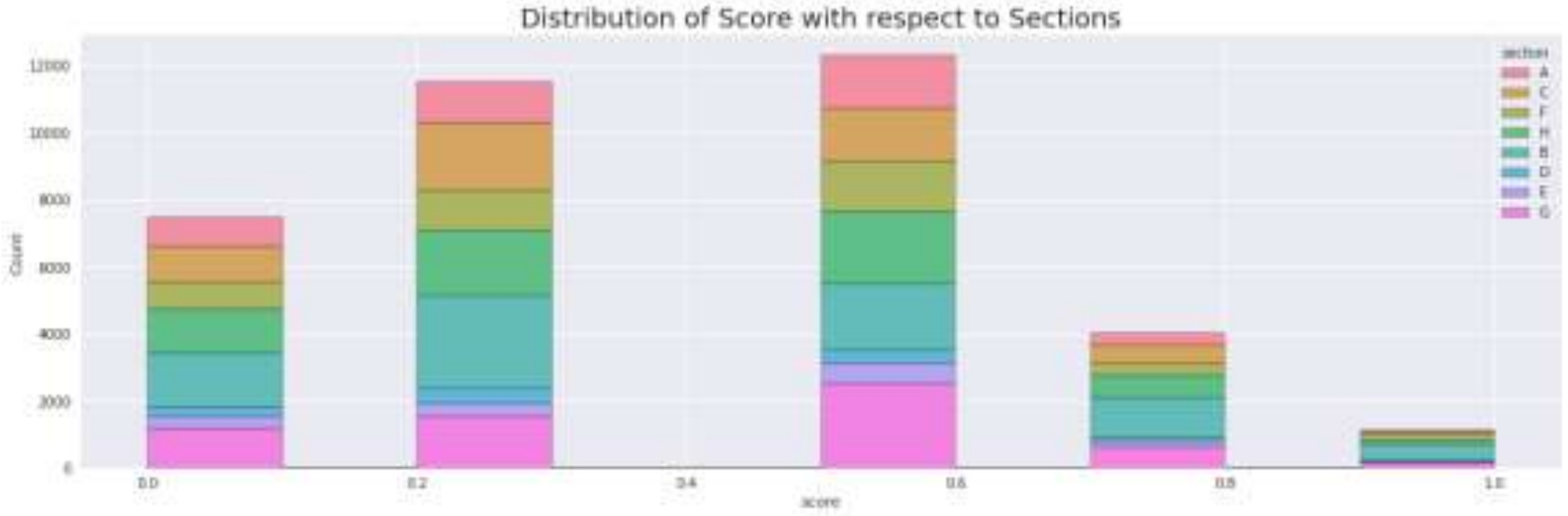

**Figure 5** Title word cloud map

## 4. Conclusion

In this study, we propose the HGM-Net deep learning framework, which integrates hierarchical contrast learning (HCL), multimodal graph attention network (M-GAT) and multi-granularity sparse attention (MSA) to significantly enhance the semantic mining capability of patent text.HCL synergistically enhances the local semantics and global topic expression through the three-tier contrast learning mechanism of words, sentences and paragraphs; M-GAT constructs patent classification coding, heterogeneous graph structure of citation and text to achieve dynamic fusion of cross-modal features; MSA adopts hierarchical sparse strategy to optimize the efficiency of long text modeling at word, phrase, sentence and paragraph granularity. Experiments show that the framework effectively solves the bottlenecks of existing methods in cross-modal alignment, long text efficiency and hierarchical semantic coherence.